%% file: main.tex
\documentclass[final,1p,times]{elsarticle}

\usepackage{amssymb}
\usepackage{amsmath}
\usepackage{hyperref}

\usepackage{multirow}
\usepackage{booktabs}
\usepackage{colortbl}
\usepackage{color,xcolor}
\usepackage{graphicx}
\usepackage{float}
\usepackage{subfig}
\usepackage{algorithm}
\usepackage{algpseudocode}

\journal{Neural Networks}

\begin{document}

\begin{frontmatter}



\title{SWIFT: Sample-Wise Instance Fine-Tuning of Time-Series Foundation Models for Forecasting}


\affiliation[1]{organization={State Key Laboratory for Novel Software Technology, Nanjing University},
                city={Nanjing},
                postcode={210008}, 
                country={China}}

\affiliation[2]{organization={School of Intelligent Science and Technology, 
Nanjing University},
                city={Suzhou},
                postcode={215163}, 
                country={China}}

\affiliation[3]{organization={Siemens Data and AI Research},
                city={Beijing},
                postcode={100102}, 
                country={China}}

\affiliation[4]{organization={Nanjing University– Siemens Joint Research Center on Industrial AI},
                city={Suzhou},
                postcode={215163}, 
                country={China}}

\author[1,2,3,4]{Yang Chen} 
\author[1,2]{Feifan Tian}
\author[1,2]{Cong Wang}
\author[3,4]{Cheng Feng}
\author[1,2,4]{Wenbin Li\corref{cor1}}
\cortext[cor1]{Corresponding author}
\ead{liwenbin@nju.edu.cn}
\author[1,2]{Yang Gao}


\begin{abstract}
While current few-shot adaptation strategies for time series forecasting foundation models are predominantly dataset-centric, the necessity of obtaining an additional dataset for the finetuning process poses a significant hurdle for adoption. To address this constraint, we propose \textit{\textbf{S}ample-\textbf{W}ise \textbf{I}nstance \textbf{F}ine-\textbf{T}uning (\textbf{SWIFT})}, a novel forecasting adaptation framework that achieves fine-grained specialization by leveraging structural priors from the target instance’s own history. This shift allows the fine-tuning process to be entirely transparent to the end-user, which significantly enhances the ease of adoption. To implement this framework, SWIFT synergizes \textit{Self-Retrieval} for historical mining and \textit{Frequency-Aware Loss} for spectral consistency, while employing \textit{Deviation-Aware Dynamic Fusion} to balance adaptation with general knowledge preservation.
Extensive experiments across four TSFM backbones and three
long-history benchmark subsets validate SWIFT, yielding average MASE
and WQL reductions of 6.41\% and 4.83\% across 24 combinations of
four backbones and six \textit{Chronos Benchmark II} datasets.
To the best of our knowledge, SWIFT is the first framework
to investigate sample-wise instance fine-tuning of time-series
foundation models, specializing the model for an individual
prediction window using only the corresponding forecasting
sample's own observed history.
\end{abstract}




\end{frontmatter}



\input{01}
\input{02}
\input{03}

\input{04}

\input{05}
\input{06}

\section*{Acknowledgement}
This work is supported in part by the National Natural Science Foundation of China (62576160, 62192783) and Nanjing University-Siemens Joint Research Center for Industrial AI.






 \bibliographystyle{elsarticle-harv} 
 \bibliography{references}




\end{document}

%% file: 01.tex
\section{Introduction}

Time Series Foundation Models (TSFMs), trained on massive corpora of diverse temporal data, have ushered in a new era of zero-shot forecasting capabilities~\citep{das2024decoder,feng2024general,khwaja2026toto,podest2026tirex}. These models excel at capturing general temporal patterns, across a wide range of domains~\citep{garza2023timegpt,gao2024units}. Despite their outstanding zero-shot capabilities, adapting TSFMs
to a specific forecasting instance remains an important
research question~\citep{dange2025tfmadapter,lee2025lightweight}.
Real-world time series exhibit heterogeneous temporal patterns
and non-stationarity~\citep{liu2022non,wan2024tcdformer,sonmezer2025canet}, with local trends and
seasonal behavior varying across sequences and over time.
These characteristics motivate investigating whether the
target instance's own observed history can provide useful
supervision for specialization.

Existing research explores adaptation through few-shot
fine-tuning~\citep{das2024context}, transfer
learning~\citep{he2023domain}, and retrieval-augmented
forecasting~\citep{ning2025ts}.
Recent studies have also investigated fine-tuning TSFMs
such as Chronos-2 on task-specific datasets, highlighting
the potential and limitations of dataset-level
adaptation~\citep{yu2026process}.
These approaches leverage target-domain training data or
relevant support examples to improve forecasting.
However, such data may not be readily available when a
forecasting request provides only the target instance's
observed history.
This constraint motivates a complementary question:
can a pretrained TSFM be specialized for an individual
prediction window using context--target pairs constructed
solely from that instance's own history?

The practical challenge is the availability and preparation
of adaptation data beyond the current forecasting sample.
While a target-domain training split may already exist in
some applications, others require additional data collection
or curation.
In streaming or privacy-restricted environments, access
restrictions may further limit the use of other
sequences~\citep{imtiaz2020privacy}.
Even when support data are available, their relevance to
the current forecasting instance must be assessed, since
distribution mismatch can limit transfer
benefits~\citep{wang2019characterizing,liu2024active}.
These constraints motivate a complementary history-only
adaptation setting.

To address this setting, we propose
\textit{Sample-Wise Instance Fine-Tuning (SWIFT)}.
A forecasting sample comprises an observed history,
a forecast origin, and a requested future prediction window.
SWIFT treats the sample's own history as an intrinsic
adaptation dataset and specializes the pretrained model
for the corresponding prediction window.
All adaptation targets are observed before the forecast origin.
To the best of our knowledge, SWIFT is the first framework
to investigate history-only sample-wise fine-tuning of TSFMs
at this prediction-window.

\begin{figure}[t]
    \centering
    \includegraphics[width=1.03\linewidth]
    {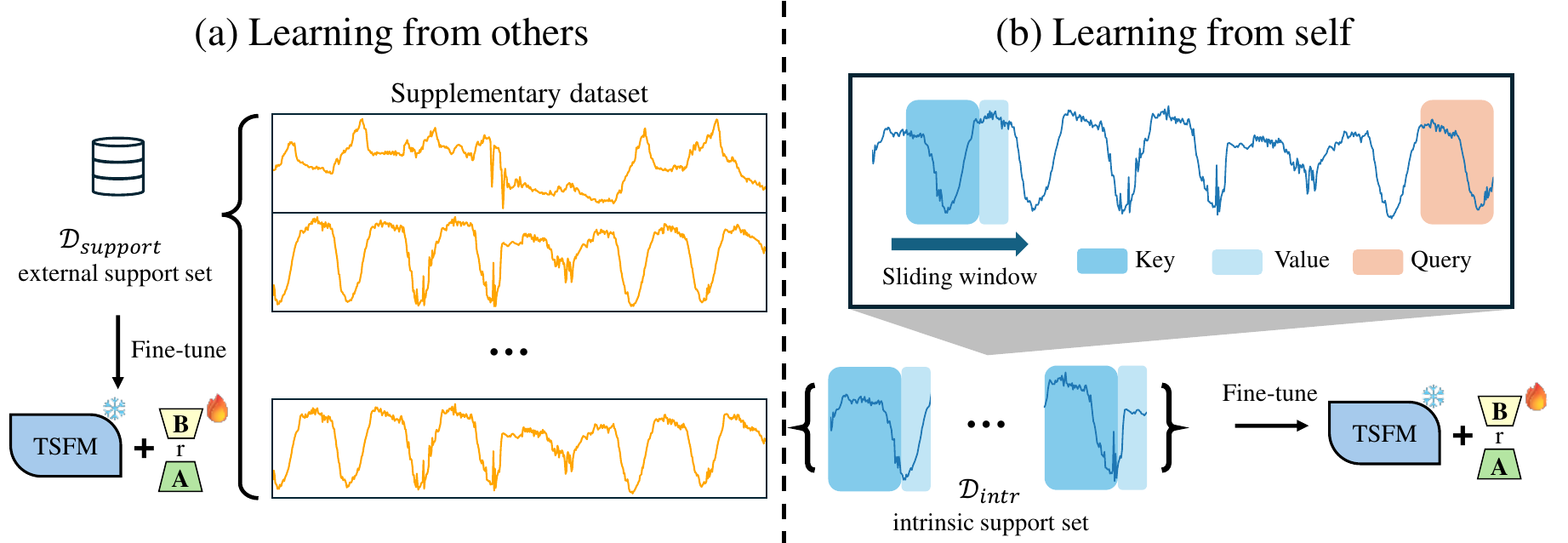}
    \caption{
    Learning from others versus learning from self.
    The illustrated support-set adaptation learns from a separately
    supplied support set
    $\mathcal{D}_{\mathrm{support}}$ of similar sequences, leveraging
    cross-instance information.
    In contrast, SWIFT constructs an intrinsic support set
    $\mathcal{D}_{\mathrm{intr}}$ by mining context--target pairs from the
    target instance's own history.
    Both approaches update only LoRA parameters, while SWIFT constructs its support set from the current sample's
    own history to specialize the model for its prediction window.
    }
    \label{fig:learning_from_self}
\end{figure}

As illustrated in Figure~\ref{fig:learning_from_self}, SWIFT fundamentally shifts the adaptation objective from ``learning from others'' to ``learning from self''. It posits that by learning from the historical behaviors inherent to the specific instance itself, we can specialize a generalist foundation model into an instance-specific expert at pre-inference adaptation time, without requiring a separate adaptation dataset or access to
other sequences, while introducing model-side computation
for historical retrieval and parameter updates.

To implement this paradigm, we address data scarcity and overfitting through three key components. First, we employ Self-Retrieval to mitigate data limitations by mining the sample's own history for relevant subsequences, unlike standard retrieval methods that query external databases ~\citep{lewis2020retrieval,ning2025ts}. Second, we refine the spectral consistency approach ~\citep{wang2024fredf} with Frequency-Aware Loss (FreAL), using adaptive weighting to ensure the model captures global dynamics rather than local fluctuations. Finally, to prevent catastrophic forgetting, we introduce a Deviation-Aware Dynamic Fusion mechanism, adapting trust-region principles ~\citep{schulman2015trust,schulman2017proximal} as an inference-time safeguard to balance the specialized expert with the robust baseline.

In summary, our specific contributions are as follows:
\begin{itemize}
    \item We pioneer sample-wise instance fine-tuning for TSFMs at the
    granularity of individual forecasting samples.
    SWIFT specializes a pretrained model for each requested
    prediction window using only the corresponding sample's
    observed history, eliminating the need for a separate
    adaptation dataset.

    \item We propose a unified framework to materialize this paradigm by synergizing three components to address core adaptation challenges: Self-Retrieval, Frequency-Aware Loss, and a Deviation-Aware Dynamic Fusion.

    \item We validate our method on curated long-history subsets of the \textit{fev-bench}, \textit{GIFT-Eval}, and \textit{Chronos Benchmark II}. Notably, SWIFT achieves a 6.99\% average WQL improvement gain on the \textit{Chronos Benchmark II} compared to the Chronos-2 zero-shot baseline.
\end{itemize}

%% file: 02.tex
\section{Related Work}

{\renewcommand{\thetable}{\Roman{table}}
\begin{table*}[t]
\centering
\caption{Comparison of SWIFT with representative adaptation paradigms for TSFMs.}
\label{tab:paradigm_comparison}
\resizebox{\linewidth}{!}{
\begin{tabular}{lcccccc}
\toprule
\textbf{Adaptation Strategy} & \textbf{Update Stage} & 
\textbf{External Data} & \textbf{Adaptation Data} & 
\textbf{Supervision Type} & \textbf{Adaptation Granularity} \\
\midrule
Zero-shot  & None & None & None & None & N/A \\
Few-shot adaptation & Offline & Typically required & Support Set & Observed targets & Dataset-level \\
TTA  & Online & None & Test Stream & Self-supervision & Stream/batch-level \\
\midrule
\textbf{SWIFT (Ours)} & \textbf{Pre-inference} & 
\textbf{None} & \textbf{Target History} & \textbf{Internal Priors} & 
\textbf{Instance-level} \\
\bottomrule
\end{tabular}
}
\end{table*}
}

\subsection{Time-Series Forecasting Models}

Classical forecasting methods, such as ARIMA
~\citep{box1968some} and related statistical models
~\citep{hyndman2018forecasting,hyndman2002state,assimakopoulos2000theta}, are commonly fitted to
individual series and can capture relatively simple temporal
dynamics. Deep forecasting models instead share parameters
across collections of series, using architectures such as RNNs
~\citep{salinas2020deepar,mienye2024recurrent}, CNNs
~\citep{luo2024moderntcn}, and Transformers
~\citep{Liu2023iTransformerIT,Zhou2023OneFA,
Zhou2022FEDformerFE,chen2024joint}. Although such global models improve
cross-series generalization, shared representations may not
fully capture instance-specific dynamics.

Recent TSFMs further scale this global modeling paradigm
through large and heterogeneous pre-training corpora
~\citep{liu2024time,wang2024timemixer++,shi2024time,
rasul2023lag}. Chronos-2~\citep{ansari2025chronos}
supports zero-shot univariate and multivariate forecasting,
while Moirai 2.0~\citep{liu2025moirai} and TimesFM
~\citep{das2024decoder} explore efficient decoder-based
architectures and large-scale pre-training. TiRex~\citep{auer2025tirexzeroshotforecastinglong} further advances large-scale temporal pre-training with strong zero-shot forecasting capabilities.These advances
provide strong zero-shot generalization, while also motivating
adaptation methods for target instances whose local dynamics differ from those captured during pre-training.

\subsection{Adaptation Paradigms for Time Series Forecasting Foundation Models}
\textbf{Zero-shot.}
Zero-shot forecasting serves as the baseline for evaluating TSFMs, utilizing pre-trained weights without parameter updates~\citep{das2024decoder,ansari2025chronos,liu2025moirai,feng2024general}. While attractive for its simplicity, zero-shot inference may exhibit reduced performance under distribution shifts, as generic representations may not fully capture target-specific dynamics.


\textbf{Few-shot adaptation.}
Few-shot fine-tuning updates a pre-trained TSFM using a support set
relevant to the target task~\citep{das2024context}. Recent strategies
such as Multi-Scale Fine-Tuning (MSFT) incorporate multi-scale temporal
modeling and outperform full fine-tuning and representative
parameter-efficient methods across multiple TSFM
backbones~\citep{qiao2026multi}. Retrieval-augmented and in-context
methods instead use examples from external time series as additional
forecasting context or demonstrations~\citep{ravuru2024agentic,
zhang2025timeraf,zhou2026stationarity,das2024context}. Although these
approaches can improve target-specific predictions, they require
target-domain or external adaptation data, and their effectiveness
depends on the quality and alignment of the selected examples. Poorly
aligned data may cause negative transfer, and fine-tuning Chronos-2 can
degrade performance on certain datasets~\citep{
xie2026posttrainingtimeseriesfoundation}.

\textbf{Test-time adaptation.}
Test-time adaptation (TTA) updates a pre-trained forecaster during
inference to track distribution shifts in an evolving test stream
~\citep{liang2025comprehensive}. Recent methods leverage progressively
revealed observations or partially observed ground truth to construct
self-supervised objectives and continually recalibrate the model under
non-stationary conditions~\citep{kim2025battling}. Although this online
adaptation can improve responsiveness to recent changes, it requires
future test-stream observations to become available and may introduce
additional waiting time. Moreover, repeated parameter updates can lead
to error accumulation, parameter drift, and unstable performance when
the incoming observations are noisy or unrepresentative.

In this work, we introduce SWIFT, a sample-wise instance fine-tuning
framework for TSFMs. A sample denotes a forecasting instance
comprising an observed history, a forecast origin, and a requested
prediction window. SWIFT specializes the pretrained model for
this window using context--target pairs drawn exclusively from
the sample's own history. Unlike dataset-level
fine-tuning, its specialization targets an individual forecasting
instance; unlike continual TTA, it completes adaptation before
forecasting using only observations available at the current origin.
Table~\ref{tab:paradigm_comparison} summarizes the representative
protocols.

%% file: 03.tex
\section{Problem Formulation}

\subsection{Preliminaries}

Let $\mathcal{X}=\{x_{1:T}\}\in\mathbb{R}^{C\times T}$ denote a
multivariate time-series instance with $C$ channels and length $T$.
Given a foundation model $f_{\theta_0}$ pre-trained on
$\mathcal{D}_{\mathrm{source}}$, zero-shot forecasting predicts the
future horizon $H$ from a context window of length $L$:
$\hat{x}_{T+1:T+H}=f_{\theta_0}(x_{T-L+1:T})$.
The ideal target-specific adaptation objective is

\begin{equation}
    \theta^*
    =
    \arg\min_{\theta}
    \mathcal{L}\!\left(
        f_\theta(x_{T-L+1:T}),
        x_{T+1:T+H}
    \right).
    \label{eq:ideal_objective}
\end{equation}

Since $x_{T+1:T+H}$ is unavailable at inference time,
Eq.~\eqref{eq:ideal_objective} is an ideal objective rather than an
executable adaptation procedure.

\subsection{External-Support Fine-Tuning}

Conventional fine-tuning assumes access to an external support set

\begin{equation}
    \mathcal{D}_{\mathrm{support}}
    =
    \left\{
        \left(x_{\mathrm{ctx}}^{(i)},x_{\mathrm{tgt}}^{(i)}\right)
    \right\}_{i=1}^{N},
\end{equation}

where each context--target pair is sampled from a support distribution
$\mathcal{P}_{\mathrm{support}}$ assumed to be aligned with the target
instance. Starting from $\theta_0$, adaptation minimizes

\begin{equation}
    \theta^*
    =
    \arg\min_{\theta}
    \mathbb{E}_{(x_{\mathrm{ctx}},x_{\mathrm{tgt}})
    \sim\mathcal{D}_{\mathrm{support}}}
    \left[
        \mathcal{L}
        \left(f_\theta(x_{\mathrm{ctx}}),x_{\mathrm{tgt}}\right)
    \right]
    \label{eq:external_adaptation}
\end{equation}

This formulation becomes unavailable when
$\mathcal{D}_{\mathrm{support}}=\varnothing$ and may be suboptimal when
the external support distribution is misaligned with the target.
Because different sequences can exhibit distinct non-stationary
dynamics, optimizing over $\mathcal{P}_{\mathrm{support}}$ does not
necessarily minimize the risk on $\mathcal{P}_{\mathrm{target}}$:

\begin{equation}
    \mathbb{E}_{x\sim\mathcal{P}_{\mathrm{support}}}
    [\mathcal{L}(f_\theta(x))]
    \neq
    \mathbb{E}_{x\sim\mathcal{P}_{\mathrm{target}}}
    [\mathcal{L}(f_\theta(x))].
    \label{eq:distribution_gap}
\end{equation}

Moreover, sufficiently long target histories can themselves contain
predictive temporal dependencies, reflected by the non-negligible
mutual information
$I(x_{1:T};x_{T+1:T+H})$.
External-support methods therefore face two practical limitations:
distribution misalignment and the unavailability or retrieval cost of
peer sequences.

\subsection{Sample-Wise Instance Fine-Tuning}

SWIFT operates under data isolation and uses only the target history
$x_{1:T}$. We assume that a sufficiently long sequence contains
recurrent temporal patterns that can provide instance-specific
supervision. SWIFT therefore constructs an intrinsic support set

\begin{equation}
    \mathcal{D}_{\mathrm{intr}}
    =
    \operatorname{MinePriors}(x_{1:T})
    =
    \left\{
        \left(x_{\mathrm{ctx}}^{(i)},x_{\mathrm{tgt}}^{(i)}\right)
    \right\}_{i=1}^{N},
    \label{eq:intrinsic_support}
\end{equation}

where $\operatorname{MinePriors}(\cdot)$ extracts informative
context--target pairs directly from the target timeline.

SWIFT then performs parameter-efficient adaptation within a constrained
subspace around the pre-trained model:

\begin{equation}
    \theta_{\mathrm{swift}}^*
    =
    \arg\min_{\theta\in\Omega(\theta_0)}
    \mathbb{E}_{(x,y)\sim\mathcal{D}_{\mathrm{intr}}}
    \left[
        \mathcal{L}_{\mathrm{task}}(f_\theta(x),y)
    \right],
    \label{eq:swift_objective}
\end{equation}

where $\Omega(\theta_0)$ denotes a parameter-efficient adaptation
subspace, such as LoRA, centered at $\theta_0$.
By learning from the target instance's own historical patterns, SWIFT
specializes the foundation model without external support data.

Realizing this formulation requires addressing three challenges:
(1) constructing a high-quality $\mathcal{D}_{\mathrm{intr}}$ from the
available history;
(2) learning structural dynamics without overfitting transient noise;
and
(3) controlling harmful over-adaptation.
The next section addresses these challenges through
\textit{Self-Retrieval}, \textit{Frequency-Aware Loss}, and
\textit{Deviation-Aware Dynamic Fusion}, respectively.

%% file: 04.tex
\section{Methodology}
\label{sec:methodology}
\begin{figure}[!t]
    \centering
    \includegraphics[width=0.9\linewidth]{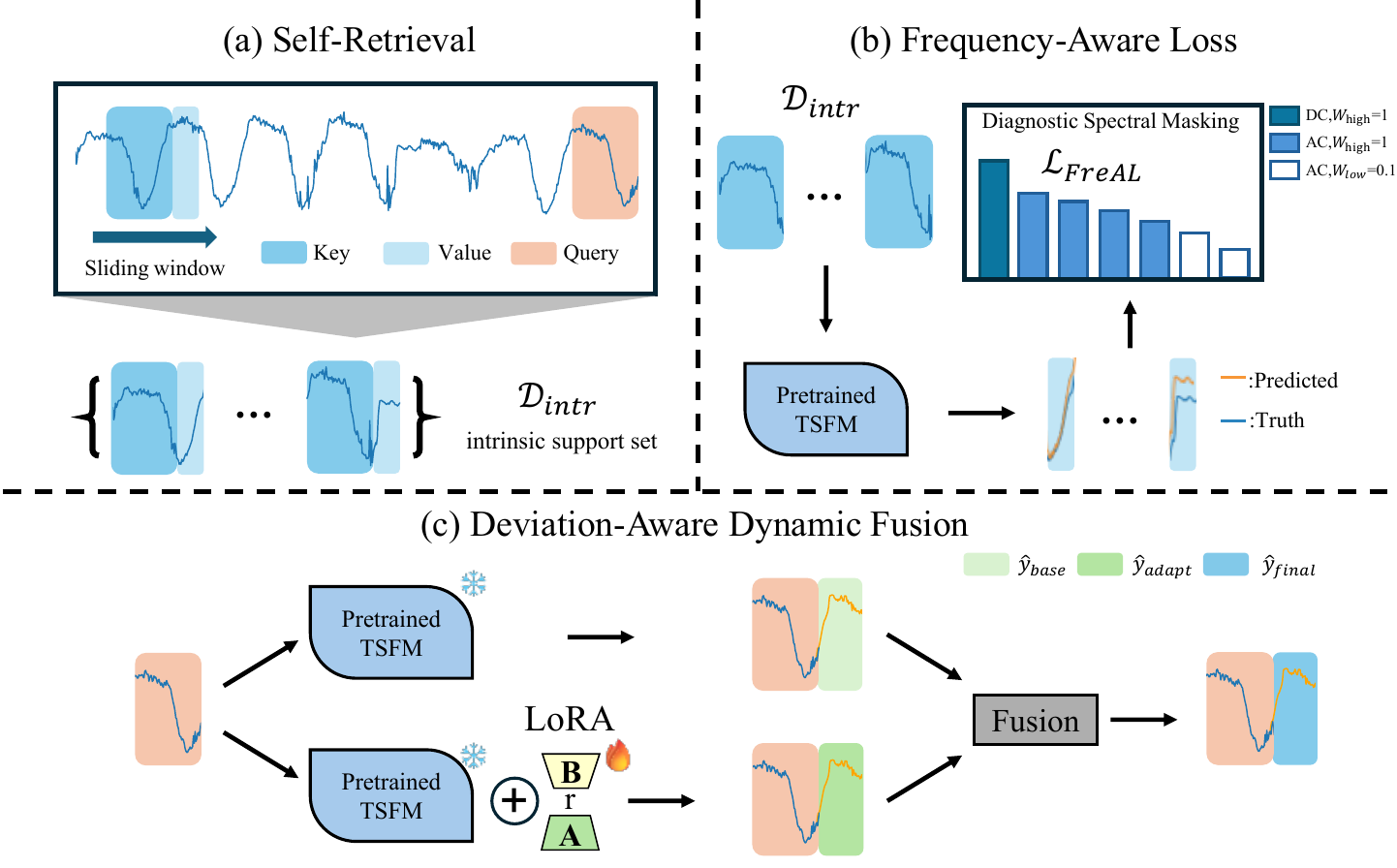}
    \caption{Overview of SWIFT. The framework consists of three stages:
    (a) \textbf{Self-Retrieval} constructs an intrinsic support set from similar historical and recent subsequences;
    (b) \textbf{Frequency-Aware Loss} adapts the model with FreAL to emphasize informative spectral discrepancies, together with an anchor regularizer that limits prediction drift;
    and (c) \textbf{Deviation-Aware Dynamic Fusion} combines adapted and zero-shot predictions to reduce harmful over-adaptation.}
    \label{fig:framework}
\end{figure}

In this section, we present the realization of the SWIFT paradigm. As illustrated in Figure~\ref{fig:framework}, SWIFT consists of three modules: (1) \textit{Self-Retrieval}, which mines structurally relevant historical and recent subsequences to construct the intrinsic support set $\mathcal{D}_{\text{intr}}$; (2) \textit{Frequency-Aware Loss} (FreAL), which guides parameter-efficient fine-tuning by emphasizing spectral discrepancies while suppressing noisy frequency components; and (3) \textit{Deviation-Aware Dynamic Fusion}, which adaptively combines the adapted and zero-shot predictions to mitigate over-adaptation. During fine-tuning, an additional prediction-space anchor regularizer constrains the adapted model's quantile forecasts toward those of the frozen foundation model.

\subsection{Self-Retrieval}
To address the challenge of data scarcity in sample-wise instance fine-tuning, we propose a Self-Retrieval mechanism.

\textbf{Query and search space construction.}
Let $x_{\text{query}}=x_{T-L_{\text{query}}+1:T}$ denote the current look-back window used as the retrieval query. The query length $L_{\text{query}}$ is determined from the dataset's known seasonality or, when such prior information is unavailable, through a spectral heuristic. In the latter case, we apply a Fast Fourier Transform (FFT) to the local context and select the period associated with the largest non-DC spectral peak.

We construct the historical search space using a sliding window of length $L_{\text{query}}$. Specifically, the window is moved along the observed history $x_{1:T}$ with stride one, producing candidate contexts
\begin{equation}
    x_{\text{cand}}^{(i)}=x_{i:i+L_{\text{query}}-1},
\end{equation}
where only windows satisfying $i+L_{\text{query}}+H-1\leq T$ are retained. This constraint ensures that each candidate context has a fully observed subsequent target window
\begin{equation}
    y_{\text{cand}}^{(i)}=x_{i+L_{\text{query}}:i+L_{\text{query}}+H-1},
\end{equation}
which can be used as supervision during adaptation. The resulting search space is
\begin{equation}
\mathcal{S}=\left\{x_{i:i+L_{\text{query}}-1}\;\middle|\;1\leq i\leq T-L_{\text{query}}-H+1\right\}.
\end{equation}

To identify candidate windows that exhibit temporal patterns similar to the query, we independently apply local Z-score normalization to each window:
\begin{equation}
    \tilde{x}=\frac{x-\mu_x}{\sigma_x+\epsilon},
\end{equation}
where $\mu_x$ and $\sigma_x$ are computed within the corresponding window. This normalization reduces the influence of level and scale differences, allowing the retrieval process to focus on temporal shape. The similarity between the query and a candidate window is then measured using the negative normalized $L_1$ distance:
\begin{equation}
s\left(x_{\text{query}},x_{\text{cand}}^{(i)}\right)
=-\frac{1}{L_{\text{query}}}\sum_{t=1}^{L_{\text{query}}}
\left|\tilde{x}_{\text{query}}^{(t)}-\tilde{x}_{\text{cand}}^{(i,t)}\right|.
\label{eq:retrieval_similarity}
\end{equation}
A larger similarity score therefore indicates a closer match between the historical candidate and the current query.

\textbf{Temperature-controlled sampling.}
Instead of greedily selecting the top-$k$ matches, which may lead to redundancy, we apply a temperature-controlled probabilistic sampling strategy to ensure diversity in the intrinsic support set. The probability of selecting the $i$-th candidate is given by
\begin{equation}
    P(i)=\operatorname{Softmax}\left(\frac{s(x_{\text{query}},x_i)}{\tau_{\text{ret}}}\right),
\end{equation}
where $\tau_{\text{ret}}$ is a temperature hyperparameter. We sample $K$ subsequences based on $P(i)$ to form the intrinsic support set $\mathcal{D}_{\text{intr}}$.

\textbf{Dual-focus augmentation.}
Finally, to ensure that the model remains responsive to recent distribution shifts and local trends, we augment $\mathcal{D}_{\text{intr}}$ with a deterministic subset of the most recent sliding windows. The retrieved and recent samples are jointly used for adaptation and are further provided as covariates during forecasting, enabling the model to combine deep historical motifs with evolving local dynamics.

\subsection{Frequency-Aware Loss (FreAL)}
\label{sec:fre_loss}
Recent work such as FreDF~\citep{wang2024fredf} has demonstrated the benefit of minimizing forecasting divergence in the frequency domain. However, existing frequency-domain objectives are primarily designed for data-rich supervised training. In our sample-wise adaptation setting, directly optimizing all frequency components may overfit high-frequency noise and transient anomalies, thereby disrupting the underlying periodic structure.

To address this issue, we propose Frequency-Aware Loss (FreAL), which adapts the spectral objective to the intrinsic support set $\mathcal{D}_{\mathrm{intr}}$ constructed by Self-Retrieval. FreAL first identifies the frequency components that contribute most to the current forecasting error, then constructs a weighted frequency-domain loss, and finally combines it with the time-domain objective.

\textbf{Diagnostic spectral weighting strategy.}
Given the prediction $\hat{y}$ and ground truth $y$, we first apply the Fast Fourier Transform to obtain their frequency representations $\hat{F}$ and $F$, and compute the absolute spectral residual $E=|\hat{F}-F|\in\mathbb{R}^{C\times L_{\mathrm{freq}}}$, where $L_{\mathrm{freq}}$ denotes the spectral length. The DC component is separated from the AC components and is always assigned a weight of $1.0$ to preserve mean-level consistency.

For the AC components, we sort their residual magnitudes in descending order, $E_{(1)}\geq E_{(2)}\geq\cdots\geq E_{(L_{\mathrm{freq}}-1)}$, and select the smallest number of components whose cumulative residual reaches a predefined threshold $\tau_{\mathrm{spec}}$:
\begin{equation}
K=\min\left\{k\;\middle|\;\frac{\sum_{i=1}^{k}E_{(i)}}{\sum_{j=1}^{L_{\mathrm{freq}}-1}E_{(j)}}\geq\tau_{\mathrm{spec}}\right\}.
\label{eq:critical_frequency_selection}
\end{equation}
The selected top-$K$ components are treated as critical frequencies and assigned $w_{\mathrm{high}}=1.0$, while the remaining components are assigned $w_{\mathrm{low}}=0.1$ to suppress less informative frequency variations. The resulting weights are then restored to their original frequency indices to form the final spectral mask $W_{\mathrm{final}}$.

Based on this mask, we define the frequency-domain loss as the weighted $L_1$ distance between the predicted and target spectra. To reduce the effect of random fluctuations during sample-wise adaptation, the loss is averaged over the batch of retrieved context--target pairs:
\begin{equation}
\mathcal{L}_{\mathrm{freq}}=\frac{1}{B}\sum_{b=1}^{B}
\left\|\left(\hat{F}_b-F_b\right)\odot W_{\mathrm{final}}^{(b)}\right\|_1,
\label{eq:frequency_loss}
\end{equation}
where $B$ denotes the number of retrieved training pairs.

Finally, FreAL combines the weighted frequency-domain loss with the time-domain forecasting loss:
\begin{equation}
\mathcal{L}_{\mathrm{FreAL}}=(1-\alpha)\mathcal{L}_{\mathrm{time}}(\hat{y},y)
+\alpha\mathcal{L}_{\mathrm{freq}}(\hat{F},F),
\label{eq:fre_loss}
\end{equation}
where $\alpha$ controls the trade-off between time-domain accuracy and spectral consistency. For quantile forecasting, $\mathcal{L}_{\mathrm{time}}$ is the pinball loss over the predicted quantiles, while the spectral objective is computed on the median forecast. Through this progressive construction, FreAL focuses adaptation on dominant spectral discrepancies while reducing overfitting to noisy tail-frequency components.

To limit prediction drift during adaptation, we introduce
an anchor loss that encourages the adapted model's outputs
to remain close to those of the frozen pretrained model:
\begin{equation}
\mathcal{L}_{\mathrm{anchor}}
=
\operatorname{MSE}\left(
\hat{y}_{\mathrm{adapt}},
\operatorname{sg}[\hat{y}_{\mathrm{base}}]
\right),
\label{eq:anchor_loss}
\end{equation}
where both predictions use the same historical support inputs,
and MSE is averaged over all quantiles and output positions.
Here, $\operatorname{sg}$ denotes stop-gradient, ensuring
that only the LoRA parameters are updated.

The complete adaptation objective is
\begin{equation}
\mathcal{L}_{\mathrm{adapt}}
=\mathcal{L}_{\mathrm{FreAL}}
+\lambda_{\mathrm{anchor}}\mathcal{L}_{\mathrm{anchor}},
\label{eq:adapt_loss}
\end{equation}
where $\lambda_{\mathrm{anchor}}\geq0$ controls the strength of regularization. Setting $\lambda_{\mathrm{anchor}}=0$ recovers adaptation with FreAL alone; increasing it penalizes departures from the frozen baseline more strongly. Anchor regularization operates during parameter updates, while deviation-aware fusion controls how the resulting forecasts are combined at inference.

\subsection{Deviation-Aware Dynamic Fusion}
A critical risk in sample-wise adaptation is ``over-adaptation'', where the model becomes too specialized to the retrieved history and fails to generalize to the immediate future, especially under sudden distribution shifts. To address this, we introduce the \textit{Deviation-Aware Dynamic Fusion} mechanism. During inference, we maintain both the frozen foundation model $f_{\theta_0}$ and the adapted expert $f_{\theta_{\mathrm{swift}}}$. We generate two sets of predictions: the robust baseline prediction $\hat{y}_{\mathrm{base}}$ and the specialized prediction $\hat{y}_{\mathrm{adapt}}$. We quantify their normalized forecast deviation as
\begin{equation}
\Delta=\frac{1}{H}\sum_{t=1}^{H}
\frac{|\hat{y}_{\mathrm{adapt}}^{(t)}-\hat{y}_{\mathrm{base}}^{(t)}|}
{|\hat{y}_{\mathrm{base}}^{(t)}|+\epsilon}.
\end{equation}
We posit that a small $\Delta$ indicates a safe adaptation where the expert refines the baseline, whereas a large $\Delta$ suggests potential overfitting or instability. We design a dynamic gating function to calculate the weight of the baseline model:
\begin{equation}
w_{\mathrm{base}}=\min\left(\lambda_{\max},\sigma\max(0,\Delta-\tau_{\mathrm{fus}})\right),
\end{equation}
where $\tau_{\mathrm{fus}}$ is a tolerance threshold and $\sigma$ is a scaling factor. The final prediction is a deviation-weighted combination:
\begin{equation}
\hat{y}_{\mathrm{final}}=(1-w_{\mathrm{base}})\hat{y}_{\mathrm{adapt}}
+w_{\mathrm{base}}\hat{y}_{\mathrm{base}}.
\end{equation}
This mechanism acts as an adaptive safety valve: it allows the specialized model to take the lead when predictions are consistent, but increases reliance on the foundation model when the adaptation deviates substantially. The same fusion weight is applied to the point and quantile forecasts. When $\lambda_{\max}<1$, the baseline contribution is capped rather than providing a complete fallback.

\subsection{Overall Procedure}
\label{sec:overall_procedure}

\begin{algorithm}[H]
\caption{SWIFT: Sample-Wise Instance Fine-Tuning}
\label{alg:swift}
\small
\begin{algorithmic}[1]
\Require Observed history $x_{1:T}$, pretrained model $f_{\theta_0}$, initial LoRA parameters $\phi_0$, context length $L$, horizon $H$, adaptation steps $S$, anchor weight $\lambda_{\mathrm{anchor}}$
\Ensure Final forecast $\hat{y}_{\mathrm{final}}$
\Statex \textbf{1. Self-Retrieval}
\State Construct query $q$ from the latest observations
\State Extract eligible historical context--target pairs
\State Score historical contexts against $q$ using Eq.~\ref{eq:retrieval_similarity}
\State Sample $K$ historical pairs using temperature-controlled top-$p$ sampling
\State Select $R$ recent context--target pairs
\State Concatenate both subsets into $\mathcal{D}_{\mathrm{intr}}$
\Statex \textbf{2. Instance-Specific Adaptation}
\State Set $\phi\gets\phi_0$ and freeze $\theta_0$
\State $(X,Y)\gets\mathcal{D}_{\mathrm{intr}}$
\For{$s=1,\ldots,S$}
    \State Predict support quantiles $\hat{Y}^{\mathcal{Q}}\gets f_{\theta_0,\phi}(X)$
    \State Obtain frozen reference quantiles $\hat{Y}^{\mathcal{Q}}_0\gets\operatorname{sg}[f_{\theta_0}(X)]$
    \State Compute $\mathcal{L}_{\mathrm{FreAL}}$ using Eq.~\ref{eq:fre_loss}
    \State Compute $\mathcal{L}_{\mathrm{anchor}}$ from $\hat{Y}^{\mathcal{Q}}$ and $\hat{Y}^{\mathcal{Q}}_0$ using Eq.~\ref{eq:anchor_loss}
    \State $\mathcal{L}_{\mathrm{adapt}}\gets\mathcal{L}_{\mathrm{FreAL}}+\lambda_{\mathrm{anchor}}\mathcal{L}_{\mathrm{anchor}}$
    \State Update only $\phi$ by minimizing $\mathcal{L}_{\mathrm{adapt}}$
\EndFor
\Statex \textbf{3. Deviation-Aware Fusion}
\State $x_{\mathrm{ctx}}\gets x_{T-L+1:T}$
\State $\hat{y}_{\mathrm{base}}\gets f_{\theta_0}(x_{\mathrm{ctx}})$
\State $\hat{y}_{\mathrm{adapt}}\gets f_{\theta_0,\phi}(x_{\mathrm{ctx}})$
\State Compute normalized forecast deviation $\Delta$
\State $w_{\mathrm{base}}\gets\min(\lambda_{\max},\sigma\max(0,\Delta-\tau_{\mathrm{fus}}))$
\State $\hat{y}_{\mathrm{final}}\gets(1-w_{\mathrm{base}})\hat{y}_{\mathrm{adapt}}+w_{\mathrm{base}}\hat{y}_{\mathrm{base}}$
\State \Return $\hat{y}_{\mathrm{final}}$
\end{algorithmic}
\end{algorithm}

Algorithm~\ref{alg:swift} summarizes SWIFT for an individual forecasting instance. Using only the observed history, SWIFT constructs an intrinsic support set, updates the LoRA parameters while freezing the pretrained backbone, and combines the adapted and zero-shot forecasts through deviation-aware fusion. The LoRA updates minimize FreAL together with the prediction-space anchor regularizer in Eq.~\ref{eq:adapt_loss}. The default setting uses 15 retrieved pairs, five recent pairs, and 15 adaptation steps. All support targets are observed before the requested forecast.

%% file: 05.tex
\section{Experiments}
\label{sec:experiments}

\subsection{Experimental Setup}
\label{sec:experimental_setup}

\textbf{Datasets and forecasting tasks.}
To support the construction of historical context--target pairs
for instance-wise adaptation, we restrict our primary evaluation
to tasks with available history length $T \geq 2048$.
Under this criterion, we select six long-history tasks from
\textit{Chronos Benchmark II}~\citep{ansari2024chronos}:
Traffic, Electricity, Exchange, ETTm, ETTh, and ERCOT.
These tasks cover transportation, electricity demand, transformer
measurements, and exchange rates, with sampling frequencies ranging
from 15 minutes to one business day.
Here, Electricity refers to the Australian Electricity dataset,
while ETTm and ETTh denote the 15-minute and hourly ETT tasks,
respectively.
We use the benchmark-specific prediction horizons: 48 steps for
Electricity, 30 steps for Exchange, and 24 steps for Traffic,
ETTm, ETTh, and ERCOT.
Within each task, all compared methods are evaluated on the same
forecasting targets and prediction horizons.

\textbf{Evaluation metrics.}
We report Mean Absolute Scaled Error (MASE) and Weighted Quantile
Loss (WQL) to assess point forecasting accuracy and probabilistic
forecasting quality, respectively.
Lower values indicate better performance for both metrics.

For a time series with observed history $y_{1:T}$ and forecast
horizon $H$, MASE is defined as
\begin{equation}
    \mathrm{MASE}
    =
    \frac{
        \frac{1}{H}\sum_{h=1}^{H}
        \left|y_{T+h}-\hat{y}_{T+h}\right|
    }{
        \frac{1}{T-m}\sum_{t=m+1}^{T}
        \left|y_t-y_{t-m}\right|
    },
\end{equation}
where $\hat{y}_{T+h}$ denotes the point forecast and $m$ is the
seasonal period specified by the benchmark.
The denominator scales the forecast error by the in-sample
mean absolute error of a seasonal naive predictor.

Probabilistic forecasts are evaluated at nine quantile levels,
$\mathcal{Q}=\{0.1,0.2,\ldots,0.9\}$.
Let $\hat{y}_{T+h}^{(q)}$ denote the predicted $q$-quantile.
The pinball loss at quantile level $q$ is
\begin{equation}
    \ell_q(y,\hat{y})
    =
    \begin{cases}
        q(y-\hat{y}), & y \geq \hat{y},\\
        (1-q)(\hat{y}-y), & y < \hat{y}.
    \end{cases}
\end{equation}
WQL averages the normalized quantile losses over
$\mathcal{Q}$:
\begin{equation}
    \mathrm{WQL}
    =
    \frac{1}{|\mathcal{Q}|}
    \sum_{q\in\mathcal{Q}}
    \frac{
        2\sum_{h=1}^{H}
        \ell_q\!\left(y_{T+h},\hat{y}_{T+h}^{(q)}\right)
    }{
        \sum_{h=1}^{H}|y_{T+h}|
    }.
\end{equation}
The above expressions describe a single forecast window;
dataset-level errors are computed following the benchmark's
aggregation protocol.

For each metric, we first compute the relative error reduction
on each dataset and then average these reductions across datasets.
Specifically, the improvement over a reference method on dataset
$d$ is
\begin{equation}
    \mathrm{Imp.}_{d}
    =
    \frac{
        E_{d}^{\mathrm{ref}}-E_{d}^{\mathrm{SWIFT}}
    }{
        E_{d}^{\mathrm{ref}}
    }
    \times 100\%,
\end{equation}
where $E_d$ denotes the corresponding dataset-level forecasting
error.
The overall improvement across $D$ datasets is
\begin{equation}
    \overline{\mathrm{Imp.}}
    =
    \frac{1}{D}\sum_{d=1}^{D}\mathrm{Imp.}_{d}.
\end{equation}
This aggregation assigns equal weight to each dataset rather
than weighting results by the number of constituent series.

To complement average relative error reductions, we report
pairwise win rates across the 24 backbone--dataset configurations.
For metric $k \in \{\mathrm{MASE},\mathrm{WQL}\}$, the win rate
of method $A$ against method $B$ is
\begin{equation}
    \mathrm{WR}_{k}(A,B)
    =
    \frac{1}{24}
    \sum_{c=1}^{24}
    \mathbb{I}\!\left[
        E_{c,k}^{A}<E_{c,k}^{B}
    \right]
    \times 100\%,
\end{equation}
where $E_{c,k}^{A}$ denotes the error of method $A$ under
configuration $c$, and $\mathbb{I}[\cdot]$ is the indicator
function.
A win requires a strictly lower error.
We also report a pooled win rate over the 48
backbone--dataset--metric comparisons:
\begin{equation}
    \mathrm{WR}_{\mathrm{pooled}}(A,B)
    =
    \frac{
        \mathrm{WR}_{\mathrm{MASE}}(A,B)
        +
        \mathrm{WR}_{\mathrm{WQL}}(A,B)
    }{2}.
\end{equation}
This pooled rate counts the two metrics separately; it does
not measure the fraction of configurations in which both
metrics improve simultaneously.

\textbf{Backbone models and inference setting.}
We evaluate SWIFT on four TSFM backbones:
Chronos-2~\citep{ansari2025chronos}, TimesFM-2.5,
Moirai-2~\citep{liu2025moirai}, and
TiRex~\citep{auer2025tirexzeroshotforecastinglong}.
The combination of four backbones and six datasets yields
24 backbone--dataset configurations.
Evaluation instances are processed independently, without
using other series as contextual demonstrations.
In particular, we set
\textit{predict\_batches\_jointly=False} for Chronos-2.
This setting separates the effects of parameter adaptation
from those of cross-sample contextual information.

\textbf{Baseline adaptation protocols.}
For each backbone, we compare SWIFT with zero-shot inference,
few-shot fine-tuning, and test-time adaptation (TTA).
The zero-shot baseline directly applies the pretrained model
without parameter updates.
The few-shot fine-tuning baseline performs dataset-level LoRA
adaptation using the full training split of the target dataset,
without an additional external support set.
Training examples are extracted independently from each
univariate series using a context length of 2048, the
dataset-specific prediction horizon, and a sliding-window
stride of 10.
Each model is fine-tuned for one epoch and subsequently
evaluated with independent per-instance inference.
For TTA, we implement the method
of~\citet{kim2025battling} on all four backbones.
The pretrained forecaster remains frozen, while a lightweight
Gated Calibration Module (GCM) calibrates its inputs and outputs.
Before forecasting each benchmark target window, the GCM
undergoes ten gradient updates on ten preceding pseudo-windows
with a unit sliding stride.
Each update minimizes mean squared error over the observed
portion of the corresponding pseudo-window target, using a
learning rate of $10^{-5}$ and weight decay of $10^{-4}$.
The calibrated model then generates the complete target horizon
in one forecasting operation.
The ground truth of the final target window is reserved
exclusively for evaluation.

These baselines represent different adaptation regimes:
dataset-level fine-tuning, rolling online calibration, and
instance-wise pre-inference adaptation.
Their adaptation data sources and update budgets therefore
differ; the comparison evaluates these protocols under
their respective information requirements.

\textbf{Implementation and runtime measurement.}
Experiments are conducted on a single NVIDIA RTX A6000 GPU
with 48\,GB of memory.
The implementation uses PyTorch and Hugging Face Transformers,
with the random seed fixed to 123.
Runtime measurements exclude model loading and dataset
downloading.
For few-shot fine-tuning, the measured computation includes
dataset-level LoRA training and subsequent forecasting;
for TTA, it includes the ten GCM updates and target forecasting.
For SWIFT, it includes support-set construction, 15 LoRA
updates, generation of the frozen and adapted forecasts,
and deviation-aware fusion.
These measurements quantify computation after the required
data are available and exclude data-collection time and
the waiting time for online observations.

\subsection{SWIFT across Diverse TSFM Backbones}
\label{sec:main_backbone_results}

\begin{table*}[t]
\centering

\setlength{\tabcolsep}{5pt}
\renewcommand{\arraystretch}{1.05}

\resizebox{\textwidth}{!}{
\begin{tabular}{l|cc|cc|cc|cc|cc|cc}
\toprule
\textbf{Method}
&
\multicolumn{2}{c}{ERCOT}
&
\multicolumn{2}{c|}{Electricity}
&
\multicolumn{2}{c|}{Exchange}
&
\multicolumn{2}{c|}{ETTm}
&
\multicolumn{2}{c|}{ETTh}
&
\multicolumn{2}{c|}{Traffic}
\\

&
MASE & WQL
&
MASE & WQL
&
MASE & WQL
&
MASE & WQL
&
MASE & WQL
&
MASE & WQL
\\
\midrule


\rowcolor{gray!15}
\textbf{Chronos-2}
&0.6275&0.0307
&1.8293&0.0117
&\underline{0.5813}&\underline{0.0456}
&0.7599&0.0736
&0.8388&0.0262
&\underline{0.8189}&0.2416
\\

+Few-shot

&0.5586&\underline{0.0284}
&1.8145&0.0115
&0.6166&0.0505
&0.7729&0.0762
&\textbf{0.7154}&\textbf{0.0245}
&0.8441&0.2558
\\

+TTA

&\underline{0.5532}&0.0294
&\underline{1.6273}&\underline{0.0109}
&0.6521&0.0661
&\underline{0.7219}&\textbf{0.0661}
&\underline{0.7609}&\underline{0.0258}
&\textbf{0.8182}&\textbf{0.2291}
\\

\textbf{+SWIFT}

&\textbf{0.5459}&\textbf{0.0278}
&\textbf{1.5938}&\textbf{0.0094}
&\textbf{0.5733}&\textbf{0.0432}
&\textbf{0.7163}&\underline{0.0686}
&0.7922&0.0260
&0.8246&\underline{0.2415}
\\

\midrule


\rowcolor{gray!15}
\textbf{TimesFM-2.5}

&0.9127&\underline{0.0414}
&1.9123&0.0119
&\underline{0.5922}&\underline{0.0532}
&0.8257&0.0814
&0.8381&\underline{0.0284}
&\underline{0.7943}&\textbf{0.2320}
\\

+Few-shot

&\underline{0.7894}&0.0437
&1.8282&0.0146
&0.6524&0.0705
&0.8301&0.1078
&0.8017&0.0324
&\textbf{0.7922}&0.2737
\\

+TTA

&\textbf{0.7634}&0.0440
&\underline{1.5318}&\underline{0.0108}
&0.5946&0.0552
&\textbf{0.7407}&\textbf{0.0682}
&\underline{0.7778}&0.0286
&0.8538&0.2711
\\

\textbf{+SWIFT}

&0.8038&\textbf{0.0381}
&\textbf{1.3516}&\textbf{0.0106}
&\textbf{0.5802}&\textbf{0.0506}
&\underline{0.8051}&\underline{0.0784}
&\textbf{0.7641}&\textbf{0.0258}
&0.7953&\underline{0.2329}
\\

\midrule


\rowcolor{gray!15}
\textbf{Moirai-2}

&\underline{0.5691}&\underline{0.0289}
&1.7889&\underline{0.0109}
&0.6310&\underline{0.0505}
&0.7584&0.0737
&\underline{0.5606}&\underline{0.0175}
&0.8214&0.2509
\\

+Few-shot

&0.5992&0.0372
&1.7758&0.0145
&0.6408&0.0690
&0.7784&0.0996
&0.6311&0.0194
&0.8506&0.2908
\\

+TTA

&\textbf{0.5275}&0.0290
&\underline{1.4545}&0.0116
&\textbf{0.5828}&0.0583
&\underline{0.7527}&\textbf{0.0700}
&0.8214&0.0265
&\textbf{0.7735}&\underline{0.2483}
\\

\textbf{+SWIFT}

&0.5692&\textbf{0.0287}
&\textbf{1.3880}&\textbf{0.0108}
&\underline{0.6309}&\textbf{0.0497}
&\textbf{0.7423}&\underline{0.0728}
&\textbf{0.5596}&\textbf{0.0171}
&\underline{0.7884}&\textbf{0.2409}
\\

\midrule


\rowcolor{gray!15}
\textbf{TiRex}

&0.9866&0.0538
&2.3844&0.0162
&0.6380&\underline{0.0540}
&0.8030&0.0807
&0.6779&\underline{0.0200}
&0.7836&\underline{0.2310}
\\

+Few-shot

&\textbf{0.8962}&\textbf{0.0425}
&\underline{1.8131}&0.0160
&\textbf{0.6130}&0.0639
&\underline{0.7786}&0.0940
&0.7440&0.0217
&0.7979&0.2751
\\

+TTA

&0.9923&0.0544
&1.8281&\underline{0.0154}
&0.7964&0.0992
&0.7854&\underline{0.0794}
&\textbf{0.6095}&0.0220
&\textbf{0.6934}&0.2741
\\

\textbf{+SWIFT}

&\underline{0.9808}&\underline{0.0524}
&\textbf{1.7481}&\textbf{0.0146}
&\underline{0.6304}&\textbf{0.0529}
&\textbf{0.7716}&\textbf{0.0786}
&\underline{0.6767}&\textbf{0.0189}
&\underline{0.7823}&\textbf{0.2209}
\\

\bottomrule
\end{tabular}
}

\caption{
SWIFT across diverse TSFM backbones.
Forecasting performance of zero-shot inference, few-shot fine-tuning,
TTA, and SWIFT on the long-history subset of
\textit{Chronos Benchmark II}. The backbone-name rows report zero-shot
results, while the remaining rows report the corresponding adaptation
methods. Lower values indicate better performance. The best and
second-best results within each backbone are marked in \textbf{bold}
and \underline{underline}, respectively.
}
\label{tab:main_results}
\end{table*}

\begin{figure*}[t]
    \centering
    \includegraphics[width=\textwidth]{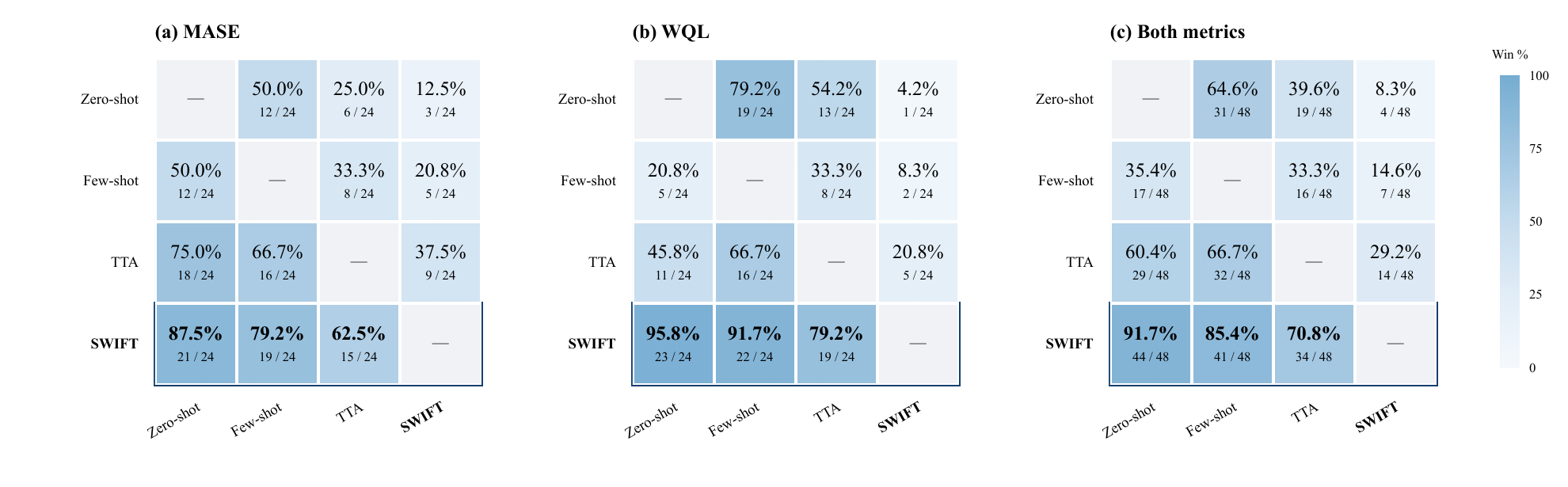}
    \caption{
        Pairwise win rates across four TSFM backbones and
        six long-history datasets from
        \textit{Chronos Benchmark II}.
        Each off-diagonal cell reports the percentage and
        count of comparisons in which the row method achieves
        a lower error than the column method.
        Panels (a) and (b) summarize 24 comparisons for MASE
        and WQL, respectively.
        Panel (c) pools the two metrics into 48 comparisons,
        rather than requiring simultaneous improvements
        in both metrics.
        Darker colors indicate higher win rates.
        Diagonal entries are omitted.
    }
    \label{fig:pairwise_win_rate}
\end{figure*}

As shown in Table~\ref{tab:main_results}, SWIFT achieves
positive average improvements over zero-shot inference
on all four backbones.
The dataset-wise relative reductions in MASE and WQL
average 6.31\% and 6.99\% for Chronos-2,
9.08\% and 6.04\% for TimesFM-2.5,
4.79\% and 1.78\% for Moirai-2, and
5.45\% and 4.50\% for TiRex, respectively.
Across all 24 backbone--dataset configurations,
the corresponding average reductions are approximately
6.41\% and 4.83\%.


\textbf{Comparison with adaptation baselines.}
Figure~\ref{fig:pairwise_win_rate} shows that SWIFT
outperforms few-shot fine-tuning in 41 of 48 comparisons,
with win rates of 79.2\% for MASE and 91.7\% for WQL,
and TTA in 34 of 48 comparisons, with corresponding
rates of 62.5\% and 79.2\%.
Its advantage is therefore more pronounced for
probabilistic forecasting.
In comparison, TTA outperforms zero-shot inference
in 75.0\% of MASE comparisons but only 45.8\% of WQL
comparisons; the corresponding rates for few-shot
fine-tuning are 50.0\% and 20.8\%.
These results show that improvements in point forecasting
do not necessarily translate into better probabilistic
forecasts.
The comparisons follow the distinct adaptation protocols
specified in Section~\ref{sec:experimental_setup}.

\textbf{Overall comparison.}
SWIFT achieves the uniquely lowest error among all four
methods in 29 of 48 comparisons (60.4\%), including
12 MASE and 17 WQL results.
Relative to zero-shot inference, it improves MASE in
21 of 24 configurations (87.5\%) and WQL in
23 of 24 configurations (95.8\%).
The four degradations occur in Moirai-2 ERCOT MASE,
Chronos-2 Traffic MASE, and TimesFM-2.5 Traffic MASE
and WQL, indicating that the benefits of adaptation
vary across backbones, datasets, and metrics.

\subsection{SWIFT across Forecasting Benchmarks}
\label{sec:cross_benchmark}

We further evaluate SWIFT with Chronos-2 on
\textit{Chronos Benchmark II}, \textit{fev-bench}
~\citep{shchur2025fev}, and \textit{GIFT-Eval}
~\citep{aksu2024gift}. The first row of
Table~\ref{tab:benchmark_avg} summarizes the Chronos-2 results from
Table~\ref{tab:main_results}, while the remaining rows extend the
evaluation to two additional benchmark collections. 


\begin{table}[t]
    \centering
    \small
    \begin{tabular}{l|c|cc}
        \toprule
        \multirow{2}{*}{\textbf{Benchmark Subset}}
        &
        \multirow{2}{*}{\textbf{\# Tasks}}
        &
        \multicolumn{2}{c}{\textbf{Relative Improvement}}
        \\
        \cmidrule(l){3-4}
        & & \textbf{WQL} & \textbf{MASE} \\
        \midrule
        \textit{Chronos Benchmark II}
        & 6 & \textbf{+6.99\%} & \textbf{+6.31\%} \\
        \textit{fev-bench}
        & 46 & \textbf{+1.66\%} & \textbf{+2.14\%} \\
        \textit{GIFT-Eval}
        & 24 & \textbf{+1.23\%} & \textbf{+1.25\%} \\
        \bottomrule
    \end{tabular}
    \caption{
        SWIFT across forecasting benchmarks.
        Average relative improvements over Chronos-2
        zero-shot inference on history-qualified tasks satisfying
        $T\geq2048$.
    }
    \label{tab:benchmark_avg}
\end{table}

SWIFT achieves positive average improvements on all three benchmark
subsets. The largest gains are observed on
\textit{Chronos Benchmark II}, with WQL and MASE reductions of 6.99\%
and 6.31\%, respectively. It also improves WQL and MASE by 1.66\% and
2.14\% on \textit{fev-bench}, and by 1.23\% and 1.25\% on
\textit{GIFT-Eval}.

\textit{fev-bench} covers diverse real-world forecasting applications,
including tasks with covariates, while \textit{GIFT-Eval} evaluates
models across varying domains, sampling frequencies, numbers of
variates, and prediction lengths~\citep{shchur2025fev,aksu2024gift}.
Despite these heterogeneous and challenging settings, SWIFT consistently
reduces both WQL and MASE, demonstrating the effectiveness of
history-only adaptation across varied forecasting scenarios. 

Overall, these results demonstrate that SWIFT generalizes across
multiple forecasting benchmarks rather than being limited to a specific
dataset collection.

\begin{table*}[t]
    \centering
    \setlength{\tabcolsep}{4.5pt}
    \renewcommand{\arraystretch}{1.05}

    \small
    \begin{tabular}{ll|cccccc|c}
        \toprule
        \textbf{Backbone}
        & \textbf{Method}
        & \textbf{Traffic}
        & \textbf{Electricity}
        & \textbf{Exchange}
        & \textbf{ETTm}
        & \textbf{ETTh}
        & \textbf{ERCOT}
        & \textbf{Avg.}
        \\
        \midrule

        \multirow{3}{*}{Chronos-2}
        & Few-shot
        & 3.708 & 55.732 & 1.437 & 16.262 & 3.722 & 36.766 & 19.604 \\
        & TTA
        & 0.661 & 3.041 & 1.609 & 1.153 & 1.007 & 1.748 & 1.537 \\
        & \textbf{SWIFT}
        & 2.748 & 5.750 & 4.712 & 2.236 & 2.953 & 4.996 & 3.899 \\
        \midrule

        \multirow{3}{*}{TimesFM-2.5}
        & Few-shot
        & 9.333 & 135.668 & 3.293 & 40.118 & 9.022 & 90.874 & 48.051 \\
        & TTA
        & 1.403 & 9.647 & 6.081 & 4.431 & 4.563 & 6.535 & 5.443 \\
        & \textbf{SWIFT}
        & 1.367 & 8.395 & 3.811 & 4.487 & 4.420 & 8.138 & 5.103 \\
        \midrule

        \multirow{3}{*}{Moirai-2}
        & Few-shot
        & 1.485 & 22.227 & 0.550 & 6.561 & 1.513 & 14.755 & 7.849 \\
        & TTA
        & 0.209 & 1.518 & 0.623 & 0.511 & 0.337 & 0.957 & 0.693 \\
        & \textbf{SWIFT}
        & 0.911 & 0.289 & 2.502 & 0.666 & 1.547 & 0.672 & 1.098 \\
        \midrule

        \multirow{3}{*}{TiRex}
        & Few-shot
        & 9.064 & 627.185 & 3.350 & 42.248 & 9.182 & 89.668 & 130.116 \\
        & TTA
        & 4.044 & 51.548 & 16.209 & 9.285 & 9.261 & 16.389 & 17.789 \\
        & \textbf{SWIFT}
        & 6.952 & 11.595 & 22.188 & 22.557 & 22.667 & 17.476 & 17.239 \\
        \bottomrule
    \end{tabular}

    \caption{
    Adaptation and forecasting latency across TSFM backbones.
    Model-side execution time in seconds for few-shot fine-tuning,
    TTA, and SWIFT on the long-history subset of
    \textit{Chronos Benchmark II}.
    Few-shot latency includes dataset-level LoRA fine-tuning and
    subsequent forecasting.
    TTA latency includes ten sequential GCM updates on preceding
    pseudo-windows and one complete target-window forecast.
    SWIFT latency includes support-set construction, 15 LoRA updates,
    base and adapted forecasting, and prediction fusion.
    Model loading, dataset downloading, and data-availability
    waiting time are excluded.
    All results are rounded to three decimal places.
    }
    \label{tab:latency_comparison}
\end{table*}

\subsection{Adaptation and Forecasting Latency}

Table~\ref{tab:latency_comparison} reports the model-side execution time
of few-shot fine-tuning, TTA, and SWIFT under their respective
adaptation protocols.
Few-shot fine-tuning includes dataset-level LoRA training
and subsequent forecasting.
TTA performs ten sequential updates of a lightweight Gated
Calibration Module (GCM) using preceding pseudo-windows,
followed by a single forecasting operation that produces
the complete target horizon.
SWIFT includes intrinsic support-set construction,
15 LoRA adaptation steps, generation of the frozen and
adapted forecasts, and deviation-aware fusion.
Across the 24 backbone--dataset combinations,
SWIFT averages 6.835 seconds, compared with 51.405 seconds
for few-shot fine-tuning and 6.365 seconds for TTA.
These results show that SWIFT requires less computation time
than the evaluated dataset-level fine-tuning protocol,
while its average runtime is close to that of TTA.

All timings measure computation after the required data
are available and exclude model loading and dataset
downloading.
TTA updates use only the observed portions of preceding
pseudo-window targets; the final target-window ground
truth is reserved exclusively for evaluation.
The unit stride specifies the movement between adaptation
windows, rather than autoregressive prediction within
the target horizon.
The ten sequential calibration updates introduce
computational overhead, which is included in the reported
TTA runtime.

TTA may additionally incur waiting
time when the observations needed for successive calibration
updates have not yet arrived.
This delay depends on the sampling interval and observation
availability and is not included in Table~\ref{tab:latency_comparison}.
When all observations required for the preceding
pseudo-window updates are already available at the forecast
origin, no additional observation-waiting delay is incurred.
SWIFT performs adaptation directly from the history available
at the current forecast origin, incurring retrieval and
optimization costs without requiring subsequent observations.
Likewise, any time needed to collect or curate the training
data for few-shot fine-tuning is excluded from the measurements.
Because the methods differ in adaptation data, trainable
components, and update budgets, the reported results
characterize the computational costs of their respective
protocols rather than a matched-budget efficiency comparison.

\subsection{Ablation Studies}
\label{sec:ablation}

\begin{table}[t]
    \centering
    \small
    \begin{tabular}{l|cc}
        \toprule
        \textbf{Setting}
        & \textbf{MASE Imp.}
        & \textbf{WQL Imp.} \\
        \midrule
        Zero-shot
        & 0.00\% & 0.00\% \\
        \textbf{Full SWIFT}
        & +6.31\% & \textbf{+6.99\%} \\
        w/o Anchor
        & +6.00\% & +6.19\% \\
        w/o Frequency
        & +4.40\% & +5.04\% \\
        w/o Fusion
        & \textbf{+6.32\%} & +5.90\% \\
        w/o RAG
        & +2.05\% & +2.38\% \\
        w/o Recent
        & +5.12\% & +4.92\% \\
        \bottomrule
    \end{tabular}
    \caption{
        Component ablation of SWIFT on
        \textit{Chronos Benchmark II}.
        MASE and WQL improvements are measured relative to
        Chronos-2 zero-shot inference.
        All adapted variants use 20 adaptation pairs and
        15 gradient steps, with all non-ablated settings held fixed.
        Dataset-wise relative improvements are computed first
        and then averaged across the six long-history datasets.
        The best result for each metric is marked in bold.
    }
    \label{tab:ablation}
\end{table}

We evaluate the contribution of individual SWIFT components using
Chronos-2 on the six long-history datasets of
\textit{Chronos Benchmark II}.
Table~\ref{tab:ablation} compares Full SWIFT with variants that
remove the anchor term, frequency-domain objective, prediction
fusion, self-retrieval, or recent-window augmentation.
The zero-shot model serves as the reference, with its relative
improvements set to zero.

All adapted variants are evaluated under strictly matched settings,
except for the component being ablated.
Specifically, they use the same pretrained checkpoint, evaluation
instances, available historical observations, context lengths,
and prediction horizons.
Each variant uses 20 adaptation pairs, a batch size of 20,
and 15 gradient update steps.
The optimizer, learning rate, LoRA configuration, initialization
procedure, and random seed are also kept unchanged.
For support-set ablations, the sample composition changes while
the total number of adaptation pairs remains fixed.
All non-ablated loss and fusion settings retain their Full SWIFT
values, without variant-specific hyperparameter tuning.
Relative improvements are computed separately for each dataset
and then averaged across the six datasets.
From the results, we make the following observations.

\begin{itemize}

    \item Self-retrieval provides the largest contribution
    among the evaluated component ablations.
    Removing self-retrieved historical windows reduces the average
    MASE improvement from 6.31\% to 2.05\% and the WQL improvement
    from 6.99\% to 2.38\%, corresponding to decreases of 4.26 and
    4.61 percentage points, respectively.
    The comparison supports the value of selecting historical
    patterns relevant to the current forecasting query.
    Nevertheless, the variant without retrieval still improves
    over zero-shot inference, indicating that adaptation remains
    useful even without the full historical selection mechanism.

    \item Recent-window augmentation complements
    self-retrieved historical patterns.
    Removing recent windows reduces the MASE and WQL improvements
    to 5.12\% and 4.92\%, respectively, compared with 6.31\% and
    6.99\% for Full SWIFT.
    These differences amount to 1.19 percentage points for MASE
    and 2.07 percentage points for WQL.
    Historical retrieval retains substantial benefits on its own,
    while incorporating recent observations further improves
    adaptation.
    This result is consistent with the dual-focus design:
    retrieved windows supply recurring historical patterns,
    whereas recent windows provide observations closer to the
    current forecast origin.

    \item Frequency-aware supervision improves both point
    and probabilistic forecasting.
    Without the frequency-domain objective, the MASE improvement
    decreases from 6.31\% to 4.40\%, while the WQL improvement
    decreases from 6.99\% to 5.04\%.
    The corresponding reductions of 1.91 and 1.95 percentage
    points indicate that the frequency-domain term provides
    benefits beyond the remaining adaptation objective.
    These results support combining time-domain supervision with
    spectral supervision when adapting the model using a limited
    historical support set.

    \item The anchor term provides a modest additional
    benefit.
    Removing the anchor term yields improvements of 6.00\% in
    MASE and 6.19\% in WQL, below the Full SWIFT results by
    0.31 and 0.80 percentage points, respectively.
    It improves both
    aggregate metrics in this evaluation.
    This suggests that the anchor constraint complements the
    adaptation objective, although these aggregate results alone
    do not establish its effect on every individual dataset.

    \item Deviation-aware fusion benefits WQL without
    improving average MASE.
    Returning the adapted forecast directly produces a MASE
    improvement of 6.32\%, slightly higher than the 6.31\%
    obtained by Full SWIFT.
    However, its WQL improvement decreases from 6.99\% to 5.90\%.
    Thus, fusion improves the aggregate WQL result by
    1.09 percentage points while introducing a negligible
    0.01-percentage-point reduction in MASE improvement.
    This comparison supports fusion as a mechanism for improving
    probabilistic forecast quality, rather than as a component
    that necessarily improves both metrics.

\end{itemize}

Overall, Full SWIFT achieves the highest average WQL improvement
and a nearly identical MASE improvement to the best ablated variant.
The matched-budget comparisons support the complementary roles
of historical retrieval, recent-window augmentation,
frequency-aware supervision, and the anchor term.
Prediction fusion provides an additional benefit for WQL,
while its effect on average MASE is minimal.
These findings support the combined framework without implying
that every component improves every metric.

%% file: 06.tex
\section{Conclusion}

In this work, we challenged the prevailing dataset-centric adaptation paradigm for Time Series Foundation Models by proposing \textit{SWIFT (Sample-Wise Instance Fine-Tuning)}, a novel approach that treats individual series as self-contained datasets to bridge the generalization-specialization gap without external corpora. SWIFT combines Self-Retrieval, Frequency-Aware Loss, and
Deviation-Aware Dynamic Fusion to construct informative supervision,
capture instance-specific dynamics, and mitigate over-adaptation.
Experiments across four TSFM backbones and three forecasting benchmarks
demonstrate consistent improvements over zero-shot inference and
competitive performance against few-shot fine-tuning and TTA without
requiring external support data or online test-stream observations.
Future work will focus on reducing the computational cost of
instance-wise adaptation.